\pdfoutput=1
\documentclass[11pt]{article}

\usepackage{ACL2023}

\usepackage{times}
\usepackage{latexsym}

\usepackage[T1]{fontenc}
\usepackage[utf8]{inputenc}

\usepackage{microtype}

\usepackage{inconsolata}

\usepackage{amsmath}
\usepackage{amsfonts}
\usepackage{graphicx}
\usepackage{subcaption}
\usepackage{tikz}
\usepackage{multirow}
\usepackage{float}
\usepackage{booktabs}
\usepackage{tcolorbox}
\usepackage{float}
\usepackage{multicol}
\usepackage{algorithm}
\usepackage{algpseudocode}
\usepackage{soul}
\usepackage{arydshln}
\usepackage{svg}

\title{Improving Semantic Control in Discrete Latent Spaces with \\Transformer Quantized Variational Autoencoders}

\author{Yingji Zhang$^{1\dagger}$,~ Danilo S. Carvalho$^{1,3}$, ~ Marco Valentino$^{2}$, \\ \textbf{~ Ian Pratt-Hartmann$^{1}$,~ Andr\'{e} Freitas$^{1,2,3}$} \\
  $^{1}$ Department of Computer Science, University of Manchester, United Kingdom\\
  $^{2}$ Idiap Research Institute, Switzerland\\
  $^{3}$ National Biomarker Centre, CRUK-MI, Univ. of Manchester, United Kingdom\\
  \texttt{$^{1}$\{firstname.lastname\}@[postgrad.]$^{\dagger}$manchester.ac.uk}
  \\ \texttt{$^{2}$\{firstname.lastname\}@idiap.ch}}

\begin{document}
\maketitle
\begin{abstract}
Achieving precise semantic control over the latent spaces of Variational AutoEncoders (VAEs) holds significant value for downstream tasks in NLP as the underlying generative mechanisms could be better localised, explained and improved upon. Recent research, however, has struggled to achieve consistent results, primarily due to the inevitable loss of semantic information in the variational bottleneck and limited control over the decoding mechanism.
To overcome these challenges, we investigate discrete latent spaces in Vector Quantized Variational AutoEncoders (VQVAEs) to improve semantic control and generation in Transformer-based VAEs. In particular, We propose T5VQVAE, a novel model that leverages the controllability of VQVAEs to guide the self-attention mechanism in T5 at the token-level, exploiting its full generalization capabilities. Experimental results indicate that T5VQVAE outperforms existing state-of-the-art VAE models, including Optimus, in terms of controllability and preservation of semantic information across different tasks such as auto-encoding of sentences and mathematical expressions, text transfer, and inference. Moreover, T5VQVAE exhibits improved inference capabilities, suggesting potential applications for downstream natural language and symbolic reasoning tasks.
\end{abstract}

\section{Introduction}
The emergence of deep generative neural networks supported by Variational AutoEncoders (VAEs) \cite{kingma2013auto} enables the localisation of syntactic and semantic properties within complex sentence latent spaces. By localising and manipulating these generative factors within the latent spaces, one can better control the properties of the textual output, enhancing performance on downstream tasks \cite{carvalho2023learning, john-etal-2019-disentangled}, and providing mechanisms for representing and disentangling syntactic and semantic features within natural language \cite{zhang2023learning, zhang2022quasi, https://doi.org/10.48550/arxiv.2109.07169}.
\begin{figure}[t]
% \begin{center}
\centering
    \includegraphics[width=\columnwidth]{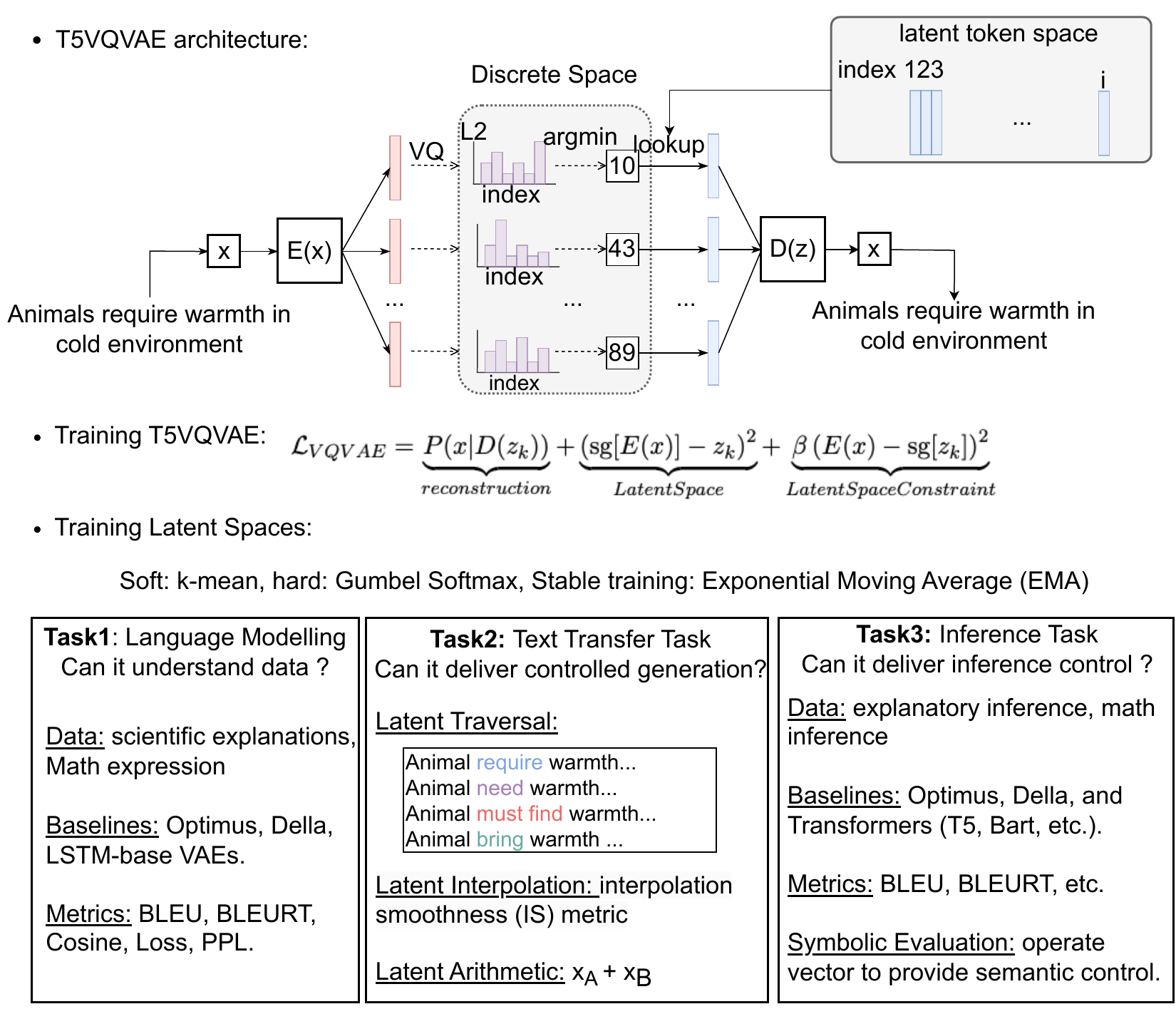}
    % \includesvg[scale=0.3]{figs/vqvae.svg}
    \caption{By controlling the token-level discrete latent space in VAEs, we aim to explicitly guide the cross-attention mechanism in T5 to improve the generation process. We focus on three challenging tasks to assess precise semantic control and inference.}
    \label{fig:overview}
 % \end{center}
\end{figure}

Recent work \cite{carvalho2023learning, zhang2022quasi, zhang2023learning} investigated controllable text generation via latent sentence geometry based on the canonical Optimus architecture (the first large pre-trained language VAE, \citet{li2020optimus}). However, the Optimus architecture brings its associated challenges since (i) the Optimus setup does not allow for a fine-grained (i.e., token-level) semantic control as sentence-level representation features are ignored by most attention heads especially in lower layers, where lexical-level semantics is captured \cite{hu-etal-2022-fuse}; (ii) the sentence bottleneck in the VAE architecture leads to inevitable information loss during inference \cite{zhang2023llamavae,zhang2023graph}.

% (i) the granular content encoding is captured at the encoder-decoder levels while broader syntactic and semantic coherent factors are captured within the latent spaces of VAE, leading to the difficulty of precise semantic control via latent spaces; (ii) the latent sentence vector, when transformed from the token level, will imply a loss of semantic information. This defines a natural functional limit to these architectures.
% In this setting, it was demonstrated that the VAE latent spaces can facilitate the control of syntactic and semantic coherence features (e.g. disentangling syntactic features and improving the alignment between predicate-argument structures and semantic roles). This points to a fundamental `division of labor' property of VAE architectures, where the granular content encoding is captured at the encoder-decoder levels while broader syntactic and semantic coherent factors are captured within the latent spaces of VAE, leading to the difficulty of precise semantic control via latent spaces. Moreover, the latent sentence vector, when transformed from the token level, will imply a loss of semantic information. This defines a natural functional limit to these architectures. 

This work concentrates on addressing these architectural limitations by aiming to minimise the information loss in the latent space and effectively control the decoder and its attention mechanism. The Vector Quantized Variational AutoEncoder (VQVAE) \cite{van2017neural}, as a discrete latent variable model, can be considered an ideal mechanism to alleviate these issues since it preserves and closely integrates both a coarse-grained continuous latent sentence space and a fine-grained latent token space that can preventinformation loss. More importantly, its latent token space can directly work on the cross-attention module \cite{vaswani2017attention} to guide the generation in seq2seq models, such as T5 \cite{raffel2020exploring}. Therefore, we hypothesise that such a mechanism can enable better generalisation and semantic control in Transformer-based VAEs.

Following these insights, we propose a novel approach named T5VQVAE, a model that leverages the controllability of VQVAE to guide the token-level self-attention mechanism during the generation process. We evaluate T5VQVAE on three challenging and diverse downstream tasks including (1) language modelling, (2) text transfer (guided text generation via the movement of latent vectors), and (3) natural language and symbolic inference tasks. An illustration of the complete model architecture and experimental setup can be found in Figure~\ref{fig:overview}. 

The overall contribution of the paper can be summarised as follows: 

\begin{enumerate}
    \item  We propose T5VQVAE, the first pre-trained language Vector-Quantised variational Autoencoder, bridging the gap between VAEs and token-level representations, improving sentence-level localisation, controllability, and generalisation under VAE architectures. The experiments reveal that the proposed model outperforms previous state-of-the-art VAE models, including Optimus \cite{li2020optimus}, on three target tasks, as well as delivering improved semantic control when compared to the previous state-of-the-art.
    \item We propose the Interpolation Smoothness (IS) metric for quantitatively evaluating sentence interpolation performance, a fundamental proxy for measuring the localisation of syntactic and semantic properties within sentence latent spaces. The experimental results indicate that T5VQVAE can lead to better interpolation paths (suggesting better interpretability and control). 
    \item Experiments on syllogistic-deductive NLI and mathematical expression derivation reveal that a quasi-symbolic behaviour may emerge in the latent space of T5VQVAE, and that the model can be explicitly controlled to achieve superior reasoning capabilities. 
\end{enumerate}
Our experimental code is available online\footnote{\url{https://github.com/SnowYJ/T5VQVAE}} to encourage future work in the field.

\section{Methodology} \label{sec:latent_props}
In this section, we first present our model, T5VQVAE, whose primary goal is to learn a latent space by reconstructing input sentences. Next, we illustrate its objective function, which consists of three parts designed to improve semantic control: reconstruction term, latent space optimization term, and encoder constraint term. Finally, we highlight the architectural advantages of T5VQVAE compared to Transformer-based VAEs.

\paragraph{Model architecture.} \citet{van2017neural} first proposed the VQVAE architecture for learning a discretised latent space of images, showing that it can alleviate the issue of \textit{posterior collapse}, in which the latent representations produced by the Encoder are ignored by the Decoder \cite{kingma2013auto}. In this work, we propose to integrate T5 encoder/decoder into the VQVAE architecture for representation learning with natural language. T5 was selected due to its consistent performance across a large range of NLP tasks and its accessibility. To cast T5 into a VQVAE model, we first establish a latent token embedding space, denoted as the codebook, represented by $z \in \mathbb{R}^{K \times I}$. Here, $K$ refers to the number of tokens in the codebook, and $I$ represents the dimensionality of each token embedding. When given a token $x$, the Encoder $E$ maps it into a vector representation, denoted as $E(x)$. Then, the nearest latent representation $z_k$ from the codebook $z$ is selected based on the $L2$ distance. The input of the cross-attention module can then be formalised as follows:
\begin{equation}
\hat{x} = \text{MultiHead}\left(D(x)W^q, z_kW^k, z_kW^v\right) \nonumber
\end{equation}
\noindent 
Here, $z_k$ is the key and value and $D(x)$, which represents the input token embedding of the decoder, is the query. $\hat{x}$ represents the reconstructed token, while $W^q$, $W^k$, and $W^v$ are trainable weights of \textbf{q}uery, \textbf{k}ey, and \textbf{v}alue.

\paragraph{Training T5VQVAE} The training of T5VQVAE can be then considered as the optimisation of three independent parts, including $D(z_k)$, $z_k$, and $E(x)$. Starting from $D$, the model can be trained by maximising the reconstruction probability $P(x|D(z_k))$ via the teach-forcing scheme. Next, the $z_k$ is optimised by minimising the $L2$ distance between $E(x)$ and $z_k$, which can be described as $(\text{sg}[E(x)] - z_k)^2$ where $\text{sg}$ is the stop gradient operation. Finally, $E(x)$ can be trained via the $L2$ distance. By ensuring that $E(x)$ can learn the latent embedding under the constraint of $R^{K \times I}$ rather than learning an embedding directly, we can guide the model to achieve better performance. A commitment weight $\beta < 1$ is used to constraint the $E$ close to $z_k$, which can be described as: $\beta (E(x) - \text{sg}[z_k])^2$. $\beta$ is set to 0.25 following the same setup as \cite{van2017neural} to preserve a behaviour consistent with their findings. The final objective function of T5VQVAE can be formalised as follows:
\[
\begin{aligned}
\mathcal{L}_{VQVAE} = \underbrace{P(x|D(z_k))}_{(1) reconstruction} + \underbrace{\left ( \text{sg}[E(x)] - z_k \right )^2}_{(2) Latent Space} \\ \nonumber
+ \underbrace{ \beta \left ( E(x) - \text{sg}[z_k] \right )^2}_{(3) Latent Space Constraint}
\end{aligned}
\]
% Where $\beta$ controls the strength of discretization of the continuous latent space. As for $E$ and $D$, we consider T5 \cite{raffel2020exploring}, a large-scale pretrained seq2seq model, to perform encoding and decoding. In the middle, the latent representations are fed into $D$ via cross-attention. Because of manipulating the cross-attention, we can efficiently control the generation of $D$. 

\begin{figure}[t]
% \begin{center}
\centering
    \includegraphics[scale=0.53]{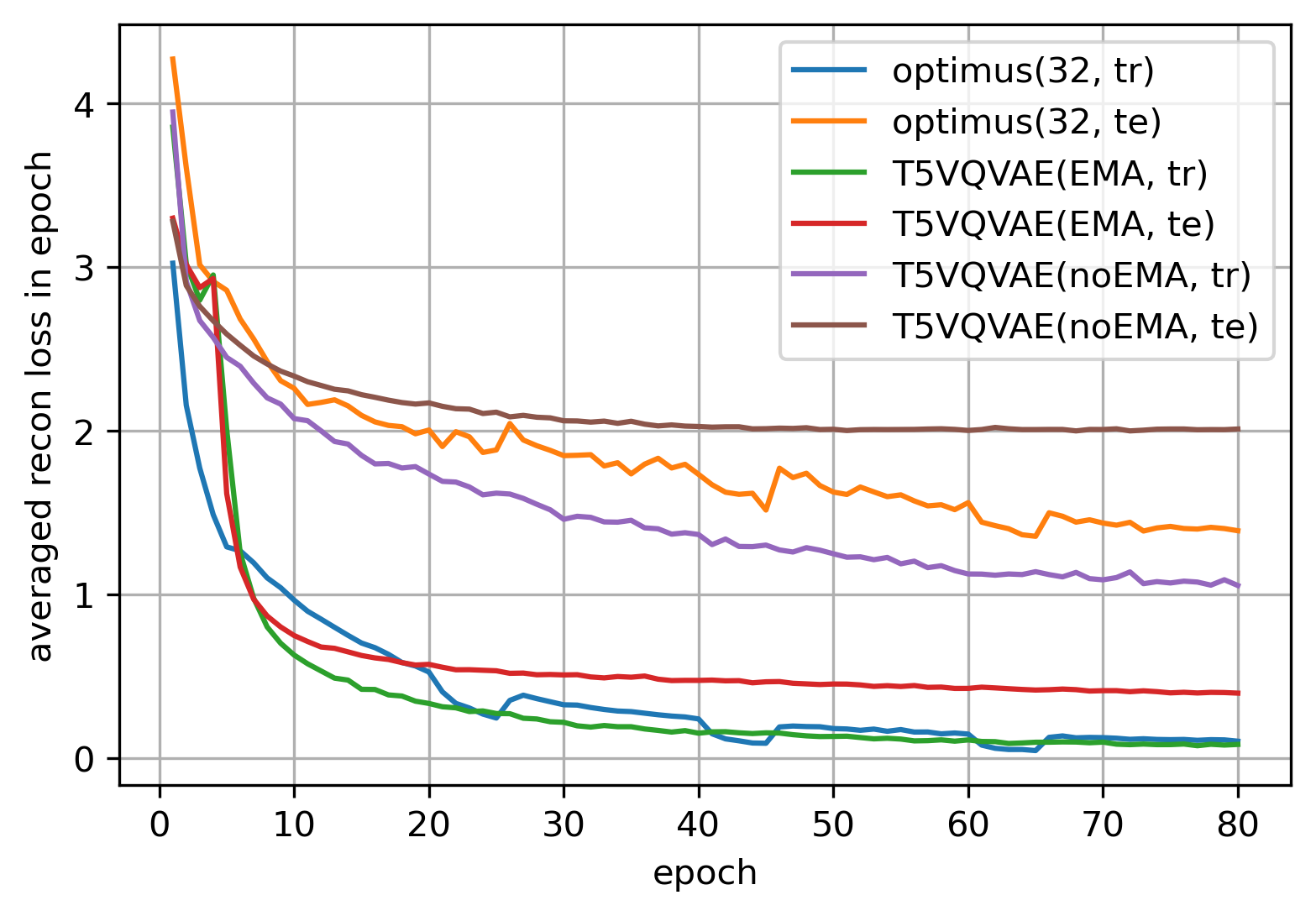}
    \caption{Loss curves of T5VQVAEs (base) with and without EMA and Optimus on the WorldTree corpus.}
    \label{fig:loss_curve}
 % \end{center}
\end{figure}

\paragraph{Training the latent space.} There are two possible strategies to update the latent space: \textit{i.} k-means and \textit{ii.} Gumbel softmax. Regarding k-means, for each token embedding $w_i$ in a sentence, it selects the nearest latent token embedding, $z_k$, to its token embedding $e^{w_i}$. This process is equivalent to classifying $e^{w_i}$ using k-means and then choosing the corresponding central point $z_k$ as the input for $D(z_k)$. This can be expressed as follows:
\[
\begin{aligned}
z_{w_i} = z_k, \enspace \text{where} \enspace k = \operatorname{argmin}_j\left\| e^{w_i} - z^j \right\|_2
\end{aligned}
\]
To improve the stability of latent space training (term 2), we adapted the Exponential Moving Average (EMA) training scheme to update $z$ \cite{roy2018theory}. 
% In detail, Let $\{E(x_{k, 1}), ..., E(x_{k, n_k})\}$ be the set of word embedding $x_{k, i}$ belonging to the $z_k$. The optimal value for $z_k$ is the average of elements in this set, which can be described as:
% \begin{align*}
% z_k = \frac{1}{n_k}\sum_{i}^{n_k} E(x_i)
% \end{align*}
% However, we cannot use this to update $z_k$ since we usually work on mini-batches. Instead, we can use EMA to update $z_k$.
% \begin{align*}
% N_k^{(t)} &:= N_k^{(t-1)} \times \lambda + n_k^{(t)} (1-\lambda) \\
% m_k^{(t)} &:= m_k^{(t-1)} \times \lambda + \sum_{i}  E(x_{k, i}) \\
% z_k & := m_k^{(t)} \div N_k^{(t)}
% \end{align*}
% Where $\lambda$ is 0.99 following the setup of \cite{van2017neural}. 
Figure \ref{fig:loss_curve} displays the training and testing loss curves of T5VQVAE with EMA or not. More details of EMA are provided in Appendix \ref{sec:set_up}.
Instead of using k-means, which performs a soft selection of the index $k$, we can utilize the Gumbel softmax trick \cite{https://doi.org/10.48550/arxiv.1611.01144} for a hard sampling of the index $k$. This trick involves sampling a noise value $g_k$ from the Gumbel distribution and then using the softmax function to normalize the output, resulting in a probability distribution. By selecting the index with the highest probability, we obtain a discrete choice. This entire process can be described as follows:
\[
\begin{aligned}
z_{w_i} &= z_k, \text{where} \\
k &= \operatorname{argmax}_k \frac{\exp (\log(t_k) + g_k)/ \tau}{\sum_{k=1}^K \exp (\log(t_k) + g_k)/ \tau} \\
\end{aligned}
\]
\noindent In this context, $t_k$ represents the probability of the $k$-th token, which can be obtained through a linear transformation before being fed into the Gumbel softmax. The parameter $\tau$ serves as a temperature hyper-parameter that controls the closeness of the new distribution to a discrete distribution. As $\tau$ approaches zero, the distribution becomes one-hot, while a non-zero value of $\tau$ leads to a more uniform distribution. In our experiments, we experienced convergence issues when using the Gumbel softmax scheme, and therefore decided to adopt the k-means mechanism which generally leads to better results.

\paragraph{Advantages of T5VQVAE.} Compared with state-of-the-art Transformer VAEs such as Optimus \cite{li2020optimus}, our model has the following architectural advantages: (i) efficient and stable latent space compression. During the training of Optimus, in fact, the KL term in ELBO is regularized cyclically \cite{fu-etal-2019-cyclical} to avoid KL vanishing and posterior collapse, which leads to an unstable training process (figure \ref{fig:loss_curve}). In contrast, T5VQVAE avoid the KL regularization term since it becomes a constant value:
\[
\begin{aligned}
\text{KL}\left(q(z_k|x)||p(z_k)\right) &= \sum_{k} q(z_k|x) \log \frac{q(z_k|x)} {p(z)} \\
&= 1 \times \log \frac{1}{1/K} = \log K
\end{aligned}
\]
where %$q(z|x)=E(x)=1$ is not a probability distribution.
the prior $p(z) = 1/K$ is a uniform distribution. (ii) Better controllability. \citet{hu-etal-2022-fuse} revealed that in Optimus \cite{li2020optimus}, the latent representation is concatenated into key and value which is more likely to be ignored by most attention heads especially in lower layers where lexical-level semantics is captured. In contrast, the latent representations of T5VQVAE are designed to act on the attention heads directly.

% To avoid the no-differentiable $E$ derived from argmin, straight-through gradient estimation \cite{bengio2013estimating} is considered for mapping the gradient from $E(x)$ to $D(z_k)$ directly by bypassing the $z_k$ and updating $z_k$ by minimizing the $L2$ distance between $E(x)$ and $z_k$, which can be described as $(\text{sg}[E(x)] - z_k)^2$ where $\text{sg}$ is the stop gradient operation. Finally, to ensure the $E$ can learn the latent embedding under the constraint of $R^{K \times I}$ rather than learn embedding directly as without latent space, the model connected by $E$ and $D$ can lead to better performance, a commitment loss is used to constraint the $E$ close to $z_k$, which can be described as: $(E(x) - \text{sg}[z_k])^2$. 

\section{Controllability Evaluation} \label{sec:eval}
Next, we put forward two metrics for quantitatively evaluating the controllability of the proposed model (T5VQVAE), which we refer to as \emph{semantic disentanglement} and \emph{interpolation smoothness}. The former evaluates the controllability from the perspective of disentanglement of semantic factors (e.g., arguments and associated semantic roles). The latter evaluates the smoothness and coherence of the latent space geometry during interpolation.

\subsection{Semantic Disentanglement} \label{sec:sem_dis} Recent studies have attempted to adapt metrics from the image domain to evaluate the semantic disentanglement of sentences \cite{zhang2022quasi, carvalho2023learning}. Semantic information in a sentence is more likely to be entangled, especially in the context of stacked multi-head self-attention models. As mentioned in \cite{zhang2022quasi, carvalho2023learning}, conceptually dense sentences are clustered according to role-content combination over the VAE latent space. Each semantic role is jointly determined by multiple dimensions rather than one single dimension. Therefore, calculating the importance of one dimension to that semantic role as a disentanglement metric is unreliable. In this work, we quantitatively evaluate the disentanglement of the semantic roles by: (1) calculating the averaged Euclidean distance between different content under that role, such as the distance between \textit{PRED-is} and \textit{PRED-are}, and (2) counting the number of different indices of the same role-content after the vector quantisation. The smaller the distance or the less the number of indices, the more concentrated the distribution of this semantic role in the latent space, indicating better disentanglement.
%However, these metrics may not be suitable for natural language evaluation since images and natural language have different distributions within the latent space. In the case of images, distinct visual features are often strongly associated with their spatial locations and possess characteristic attributes, such as the nose and eyes in facial images. 
%Consequently, the latent space can more naturally disentangle these visual concepts, with different dimensions controlling different visual aspects. In this scenario, the importance of a particular dimension for a specific visual concept can be considered a viable disentanglement metric.

 % (Euclidean space) where each dimension follows an independent standard Gaussian distribution
\subsection{Interpolation Smoothness} \label{sec:interpolate_smooth} Interpolation is a standard process for evaluating the geometric properties of a latent space in both image and language domains \cite{li2020optimus, https://doi.org/10.48550/arxiv.2106.09016}. It aims to generate a sequence of sentences following a spatial trajectory from source to target via latent arithmetics. For example, in the VAE latent space, the interpolation path can be described as $ z_t = z_1 \cdot (1 - t) + z_2 \cdot t  $  with $ t $ increased from $ 0 $ to $ 1 $ by a step size of $ 0.1 $ where $ z_1 $ and $ z_2 $ represent latent vectors of source and target sentences, respectively. In this case, each intermediate output $D(z_t)$ should change fewer semantic concepts at each step if the latent space is smooth and regular. In this work, we employ a similar strategy, however follow the more granular token level within the VQVAE. We directly manipulate the interpolation within the latent token space. At each step $t$, we obtain the intermediate latent token embedding $z^{w_i}_t$ within a sentence by calculating the weighted minimal distance between its preceding token embedding $z^{w_i}_{t-0.1}$ and the target token embeddings $z^{w_i}_2$. This process can be described as follows:
\[
\begin{aligned}
z_1^{w_i} &= e^{k_1}, z_2^{w_i} = e^{k_2}, \text{where} \enspace i = [1, ..., L] \\ \nonumber
z^{w_i}_t &= z^k, \text{where} \\
k &= \operatorname{argmin}_j \enspace (1-t) \times \left\| z_{t-0.1}^{w_i} - z^j \right\|_2 \\
&+ t \times \left\| z_2^{w_i} - z^j \right\|_2 \\
s_t &= \left[z^{w_1}_t; \dots; z^{w_L}_t\right]
\end{aligned}
\]
where $s_t$ represents the sentence embeddings at step $t$. The final generated sentence can be decoded as $s_t = D(s_t)$. Once we have obtained the interpolation path, we introduce the interpolation smoothness (IS) metric to quantitatively evaluate its smoothness. This metric involves calculating the aligned semantic distance between the source and the target (referred to as the ideal semantic distance). Subsequently, we calculate the sum of the aligned semantic distances between each pair of adjacent sentences in the path (referred to as the actual semantic distance). Finally, by dividing the ideal semantic distance by the actual semantic distance, we obtain a measure of smoothness. If the result is 1, it indicates that the actual path aligns perfectly with the ideal path, suggesting better geometric properties. Conversely, it suggests a less coherent transformation path, indicating poorer geometric properties. The metric is defined as follows:
\[
\begin{aligned}
\text{IS} = \mathbb{E}_{(s_0, ..., s_T) \sim P} \frac{\delta(\text{align}(s_0, s_T))}{\sum^T_{t=0} \delta(\text{align}(s_t, s_{t+0.1}))}
\end{aligned}
\]
\noindent where $\delta$ and $\text{align}$ are sentence similarity and alignment functions, respectively. In this experiment, sentence similarity and alignment are performed via Word Mover’s Distance \cite{zhao-etal-2019-moverscore} since it can softly perform the semantic alignment.

% In more general, the semantic alignment via automatic semantic role labelling tool \cite{Gardner2017ADS} \footnote{\url{https://allenai.org/allennlp}}, and $\delta$ could be any kinds of recent pretrained large language models, such as sentenceT5 \cite{https://doi.org/10.48550/arxiv.2108.08877}.

% , sentenceBERT \cite{reimers2019sentence}, or even out-of-shelf pretrained models, such as word2vec \cite{mikolov2013efficient} and GolVe \cite{pennington-etal-2014-glove}.

% This process can be described as:
% \[
% \begin{aligned}
% \delta(s_0, s_T) = \sum_{(t^0_{x:y}, t^T_{x_1:y_1})} \left\| z(E(t^0_{x:y})), z(E(t^T_{x_1:y_1})) \right\|
% \end{aligned}
% \]

\section{Experiments} \label{sec:empirical}

% As displayed in Figure \ref{fig:overview}, we evaluate the controllablilty of T5VQVAE on three progressively deeper tasks with the target of precise semantic control.

\subsection{AutoEncoding Task}
\paragraph{{Pre-training Data.}} In this work, we focus on the use of conceptually dense explanatory sentences \cite{https://doi.org/10.48550/arxiv.2104.08661}  and mathematical latex expressions \cite{meadows2023symbolic} to evaluate model performance. The rationale behind this choice is that (1) explanatory sentences provide a semantically challenging yet sufficiently well-scoped scenario to evaluate the syntactic and semantic organisation of the space \cite{thayaparan2020survey,valentino2022hybrid,valentino-etal-2022-case}; (2) mathematical expressions follow a well-defined syntactic structure and set of symbolic rules that are notoriously difficult for neural models \cite{meadows2023generating}. Moreover, the set of rules applicable to a mathematical expression fully determines its semantics, allowing for an in-depth inspection and analysis of the precision and level of generalisation achieved by the models \cite{welleck2022symbolic,valentino2023multi}. Firstly, we conduct a pre-training phase, evaluating the performance of T5VQVAE in reconstructing scientific explanatory sentences from WorldTree \cite{jansen2018worldtree} and mathematical latex expressions from the dataset proposed by \citet{meadows2023symbolic}. 

% More details on the datasets are provided in the Appendix \ref{sec:set_up}.
% two datasets commonly employed in the explanation-based NLI tasks: WorldTree \cite{jansen2018worldtree} and EntailmentBank \cite{https://doi.org/10.48550/arxiv.2104.08661}. Given that there is a partial overlap in the data from these two datasets, we specifically choose non-repetitive explanations.

\paragraph{Baselines.} We consider both \textit{small} and \textit{base} versions of pretrained T5 to initialise the T5VQVAE, where the codebook size is 10000. The effect of different codebook sizes on its performance and the optimal point within the architecture (different hidden layers of the encoder) to learn the codebook are reported in Table \ref{tab:diff_codebook}. As for the large VAE model, we consider Optimus with random initial weights and pre-trained weights \cite{li2020optimus} and Della \cite{hu-etal-2022-fuse}. We chose two different latent dimension sizes (32 and 768) for both of them. Moreover, we also select several LSTM language autoencoders (AE), including denoising AE (\citet{10.1145/1390156.1390294}, DAE), $\beta$-VAE \cite{Higgins2016betaVAELB}, adversarial AE (\citet{makhzani2016adversarial}, AAE), label adversarial AE (\citet{rubenstein2018latent}, LAAE), and denoising adversarial autoencoder (\citet{shen2020educating}, DAAE). Additional details on the training setup are provided in Appendix \ref{sec:set_up}. The full source code of the experimental pipeline is available at an anonymised link for reproducibility purposes.
% \begin{center}
\begin{table}[ht!]
\scriptsize
\setlength\tabcolsep{2.5pt}
\resizebox{7.8cm}{!}{
\small
\centering
\renewcommand\arraystretch{1}
\begin{tabular}{llllll} %{lp{1.5cm}p{1.5cm}p{1.5cm}p{1.5cm}p{1.5cm}} %
\toprule
\multicolumn{6}{c}{\textit{Explanatory sentences}} \\
Evaluation Metrics & BLEU & BLEURT & Cosine & Loss $\downarrow$ & PPL $\downarrow$ \\ \hline
DAE(768) & \textbf{0.74} & \textbf{0.03} & \textbf{0.91} & \textbf{1.63} & \textbf{5.10} \\ % 
AAE(768) & 0.35 & -0.95 & 0.80 & 3.35 & 28.50 \\ % \cite{makhzani2016adversarial} 
LAAE(768) & 0.26& -1.07& 0.78& 3.71 & 40.85 \\ % \cite{rubenstein2018latent} 
DAAE(768) & 0.22& -1.26& 0.76& 4.00 & 54.59 \\ % \citet{shen2020educating} 
$\beta$-VAE(768) & 0.06& -1.14& 0.77& 3.69 & 40.04 \\ \hdashline % \cite{Higgins2016betaVAELB} 
Optimus(32, rand) & 0.54 & 0.14 & 0.92 & 1.08 & 2.94 \\
Optimus(32, pre) & 0.61 & 0.29 & 0.93 & 0.86 & 2.36 \\
Optimus(768, rand) & 0.49 & -0.04 & 0.90 & 1.32 & 3.74 \\
Optimus(768, pre) & 0.68 & 0.48 & 0.95 & 0.65 & 1.91 \\
DELLA(32, rand) & 0.71 & 0.06 & 0.92 & 0.50 & 1.65 \\ % \cite{hu-etal-2022-fuse}
DELLA(768, rand) & 0.72 & 0.21 & 0.95 & \textbf{\textcolor{blue}{0.41}} & \textcolor{blue}{\textbf{1.51}} \\
% \multicolumn{6}{c}{\textit{Our approach}} \\
T5VQVAE(small, soft) & 0.81 & \textcolor{blue}{\textbf{0.62}} & \textcolor{blue}{\textbf{0.97}} & 0.46 & 1.58 \\
T5VQVAE(base, soft) & \textcolor{blue}{\textbf{0.82}} & \textcolor{blue}{\textbf{0.62}} & \textcolor{blue}{\textbf{0.97}} & 0.75 & 2.11 \\ \hline \hline
\multicolumn{6}{c}{\textit{Mathematical expressions}} \\ 
Evaluation Datasets & EVAL & VAR & EASY & EQ & LEN \\ \hline
DAE(768) & \textbf{0.94} & \textbf{0.50} & \textbf{0.80} & \textbf{0.74} & \textbf{0.58} \\
AAE(768) & 0.41 & 0.41 & 0.39 & 0.41 & 0.52 \\
LAAE(768) & 0.41 & 0.45 & 0.39 & 0.39 & 0.49 \\ % \cite{rubenstein2018latent} 
DAAE(768) & 0.38 & 0.48 & 0.35 & 0.38 & 0.49 \\ % \citet{shen2020educating} 
$\beta$-VAE(768) & 0.39 & 0.48 & 0.37 & 0.39 & 0.50 \\ \hdashline % \cite{Higgins2016betaVAELB}
Optimus(32, rand) & 0.95 & 0.59 & 0.75 & 0.71 & 0.50 \\
Optimus(768, rand) & 0.96 & 0.61 & 0.79 & 0.75 & 0.54 \\
DELLA(32, rand) & \textcolor{blue}{\textbf{1.00}} & 0.55 & 0.89 & 0.72 & 0.63 \\
DELLA(768, rand) & \textcolor{blue}{\textbf{1.00}} & 0.55 & 0.93 & 0.79 & 0.64 \\
% \multicolumn{6}{c}{\textit{Our approach}} \\
T5VQVAE(small, soft) & 0.97 & \textcolor{blue}{\textbf{0.65}} & \textcolor{blue}{\textbf{0.95}} & \textcolor{blue}{\textbf{0.90}} & \textcolor{blue}{\textbf{0.69}} \\
T5VQVAE(base, soft) & 0.98 & 0.62 & \textcolor{blue}{\textbf{0.95}} & 0.85 & 0.68
% \multicolumn{6}{c}{\textit{\textit{T5VQVAE(small, soft) with different codebook sizes}}} \\
% T5VQVAE(small, soft, 02000) & 0.73 & 0.21 & 0.93 & 0.79 & 2.20 \\
% T5VQVAE(small, soft, 06000) & 0.79 & 0.45 & 0.95 & 0.61 & 1.84 \\
% T5VQVAE(small, soft, 10000) & 0.81 & 0.62 & 0.97 & 0.46 & 1.58 \\
% T5VQVAE(small, soft, 14000) & 0.82 & 0.62 & 0.97 & 0.42 & 1.52 \\
% T5VQVAE(small, soft, 18000) & 0.83 & 0.64 & 0.97 & 0.38 & 1.46 \\
% T5VQVAE(small, soft, 22000) & 0.83 & 0.67 & 0.97 & 0.34 & 1.40 \\ \hline \hline
\\ \toprule
\end{tabular}
}
\caption{AutoEncoding task evaluation on the test set (soft: k-means). The highest scores of large VAE models and LSTM-based VAE models are highlighted in blue and in bold separately.} \label{tab:semantic_similarity_autoencoding}
\end{table}
% \end{center}
\paragraph{Quantitative Evaluation.} As for modelling explanatory sentences, we quantitatively evaluate the performance of the models using five metrics, including BLEU \cite{Papineni02bleu:a}, BLEURT \cite{https://doi.org/10.48550/arxiv.2004.04696}, cosine similarity from pre-trained sentence T5 \cite{https://doi.org/10.48550/arxiv.2108.08877}, cross-entropy (Loss), and perplexity (PPL). As for modelling mathematical expressions, we use BLEU to evaluate the robustness of models on the 5 test sets proposed by \citet{meadows2023symbolic}, one designed to assess in-distribution performance, and four designed to assess out-of-distribution generalisation. Here we provide a full characterisation of the test sets: (1) EVAL: contains mathematical statements following the same distribution of the training set (like $U + cos{(n)}$), including expressions with similar lengths and set of symbols (2) VAR: full mathematical statements with variable perturbations (like $U + cos{(beta)}$), designed to test the robustness of the models when dealing with expressions containing variables never seen during training; (3) EASY: simpler mathematical expressions with a lower number of variables, designed to test length generalisation (like $cos{(n)}$), (4) EQ: full mathematical statements with equality insertions (like $E = U + cos{(n)}$), designed to test the behaviour of the model on equivalent mathematical expressions with minimal perturbations (5) LEN: mathematical statements with a higher number of variables (like $U + cos{(n)}) + A + B$), designed to test generalisation on more complex expressions. 

As shown in Table \ref{tab:semantic_similarity_autoencoding}, the highest scores for large VAE models and LSTM-based VAE models are highlighted in blue and bold, respectively. Among them, T5VQVAEs with the k-means scheme outperforms Optimus and LSTM-based VAEs in both corpora and compared with Della, it can deliver better generation and generalization. 
We provide examples with low BLEURT scores in Appendix \ref{sec:autoencode_appendix} 
% and the optimal point within the architecture (different hidden layer of the encoder) to learn the codebook in Table \ref{tab:diff_codebook}.

% Furthermore, we investigate the optimal point within the architecture to learn the codebook. As demonstrated in the lower part of Table \ref{tab:semantic_similarity_autoencoding}, T5VQVAE demonstrates better performance when the codebook is learned at the end of the Encoder. This observation suggests that cross-attention plays a crucial role in vector quantisation (VQ) learning. 
Next, we quantitatively evaluate the disentanglement of T5VQVAE following the semantic disentanglement reference metric \ref{sec:sem_dis}. As displayed in Table \ref{tab:stat_disentanglement}, the number of central points for \textit{PRED} is higher than the remaining role-content, being 24 in \textit{PRED-is} and 6 in \textit{PRED-are}. This indicates that the semantic information of \textit{PRED} is more widely distributed in the latent space when compared to other roles. This behaviour might be attributed to the fact that the aforementioned predicates are widely used across sentences in the corpus. The full visualisation of the semantic disentanglement achieved by T5VQVAE is provided in Figure \ref{fig:tsne_big}.
\begin{table}[t]
\scriptsize
\setlength\tabcolsep{2.5pt}
\small
\centering
% \resizebox{7.8cm}{!}{
\renewcommand\arraystretch{1}
\begin{tabular}{lllll}
\toprule
Role-content & NUM centers & AVG dis & MAX dis & MIN dis \\ \hline
ARG0-animal & 3 & 0.28 & 0.52 & 0.35 \\ 
ARG1-animal & 3 & 0.28 & 0.52 & 0.35 \\
ARG2-animal & 4 & 0.33 & 0.55 & 0.35 \\
PRED-is & 24 & 0.60 & 1.08 & 0.22 \\
PRED-are & 6 & 0.31 & 0.64 & 0.21 \\
MOD-can & 5 & 0.40 & 0.82 & 0.28 \\ 
NEG-not & 2 & 0.25 & 0.51 & 0.51 \\ \toprule
\end{tabular}
% }
\caption{Semantic role disentanglement.} \label{tab:stat_disentanglement}
\end{table}
% \paragraph{Qualitative evaluation} In addition, we manually evaluate its performance and show the common issues in the AutoEncoding setup. (1) repetition: some explanations that describe the synonym are suffered from information loss. E.g., the prediction is \textit{a tool is a kind of tool} where the golden is \textit{a tool is a kind of invention}. (2) wrong numerical token: cannot precisely reconstruct the numerical token. E.g., \textit{a human sperm cell has ( 25 ; twenty - three ) chromosomes} compared with the golden: \textit{a human sperm cell has ( 23 ; twenty - three chromosomes )}. More examples with low BLEURT scores are provided in Appendix \ref{sec:autoencode_appendix}.
% \footnote{\url{https://allenai.org/allennlp}}

% \begin{figure}[ht!] 
%     \includegraphics[scale=0.33]{figs/Picture.png} 
%     \caption{t-SNE plot of the T5VQVAE latent space. Left: same role-content(\textit{PRED-is}, \textit{ARG2-animal}). Middle: different role-content(\textit{ARG0-PRED-ARG1}, \textit{ARG1-PRED-ARG2}). Right: different roles with same content (\textit{ARG0, 1, 2 - animal}, \textit{ARG0, 1, 2 - water}). A big-size visualization is provided in Figure \ref{fig:tsne_big}.}
%     \label{fig:tsne} 
% \end{figure} 

\subsection{Text Transfer Task} 
Next, we investigate the controllability of T5VQVAE by manipulating the latent space via geometric transformations. This is referred to as the Text Transfer task. We compare the performance of T5VQVAE (base, soft) and Optimus (32, pretrain) - both trained in the AutoEncoding task - as baselines. We evaluate the latent space using latent traversal, interpolation, and vector arithmetics.

\paragraph{Latent Traversal.} The traversal is inspired by the image domain, only changing the feature interpretation \citep{higgins2016beta,kim2018disentangling}. Specifically, if the vector projection within the latent space can be modified when traversing (re-sampling) one dimension, the output should only change well-defined semantic features corresponding to that dimension. In this experiment, the traversal is set up from a starting sentence. As illustrated in Table \ref{tab:trav_examples}, the T5VQVAE can provide localised semantic control by operating the discrete latent space. Different dimensions in the discrete sentence space can control different parts of the sentence. The traversal for Optimus is provided in Appendix \ref{sec:text_transfer_appendix}. 
% traversals results
\begin{table*}[ht!]
\centering
\begin{tcolorbox}[fontupper=\small, fontlower=\small] %, middle=0.3cm
\begin{multicols}{2}
    \underline{\textbf{an animal requires warmth in cold environments}} \\ \\
    dim0: \textcolor{blue}{an} animal requires warmth in cold environments \\
    dim0: \textcolor{blue}{a} animal requires warmth in cold environments \\
    dim0: \textcolor{blue}{the} animal requires warmth in cold environments \\
    
    % dim1: an \textcolor{blue}{animal} requires warmth in cold environments \\
    dim1: an \textcolor{blue}{organism} requires warmth in cold environments \\
    dim1: an \textcolor{blue}{animal} requires warmth in cold environments \\
    dim1: an \textcolor{blue}{object} requires warmth in cold environments \\
     
    dim2: an animal \textcolor{blue}{needs} warmth in cold environments \\
    dim2: an animal \textcolor{blue}{must find} warmth in cold environments \\ 
    dim2: an animal \textcolor{blue}{brings} warmth in cold environments \\
    dim2: an animal \textcolor{blue}{wants} warmth in cold environments
   \columnbreak
    \\ \\ \\
    dim4: an animal requires warmth \textcolor{blue}{during} cold temperatures \\
    % dim4: an animal requires warmth \textcolor{blue}{among} cold environments \\
    dim4: an animal requires warmth \textcolor{blue}{in} cold environments \\
    dim4: an animal requires warmth \textcolor{blue}{to} cold environments \\
    
    % dim5: an animal requires warmth in cold environments \\
    dim5: an animal requires warmth in temperatures \\
    dim5: an animal requires warmth in \textcolor{blue}{warm} environments \\ 
    dim5: an animal requires warmth in \textcolor{blue}{a warm} environment \\
    
    % dim6: an animal requires warmth in cold \textcolor{blue}{environments} \\
    dim6: an animal requires warmth in cold \textcolor{blue}{temperatures} \\
    dim6: an animal requires warmth in cold \textcolor{blue}{climates} \\
    dim6: an animal requires warmth in cold \textcolor{blue}{systems}
\end{multicols}
\end{tcolorbox}
\caption{T5VQVAE(base): traversals showing \textcolor{blue}{controlled} semantic concepts in explanations. We also provide the traversal of Optimus latent space for comparison in Table \ref{tab:appendix_traversal}.}
\label{tab:trav_examples}
\end{table*}

\paragraph{Latent Interpolation.} As described in section \ref{sec:interpolate_smooth}, interpolation aims to generate a sequence of sentences from source to target via latent vector arithmetic. An ideal interpolation should lead to reasonable semantic controls at each step. In Table \ref{tab:interpolation}, we can observe that compared with Optimus's interpolation (bottom) where the semantics are changed redundantly, e.g., from \textit{some birds} to \textit{some species mammals} to \textit{most birds} and from \textit{have} to \textit{don't have} to \textit{have}, T5VQVAE (top) leads to a more reasonable (coherent/smoother) pathway. E.g., from \textit{speckled brown color} to \textit{speckled brown feathers} to \textit{speckled wings} to \textit{wings}. Additional examples are provided in Appendix \ref{sec:text_transfer_appendix}.
% traversals results
\begin{table}[t]
\centering
\begin{tcolorbox}[fontupper=\small, fontlower=\small] %, middle=0.3cm
\underline{\textbf{Source: some birds have a speckled brown color}} \\ \\
% some birds have a speckled brown color \\
% some birds live in an arctic environment \\
% some arctic animals live in a speckled green environment \\
% \textcolor{blue}{arctic animals live in an arctic environment}\\ \\
1. \textcolor{blue}{some birds} \ul{have} \textcolor{orange}{a speckled brown color} \\
2. \textcolor{blue}{some birds} \ul{do not have} \textcolor{orange}{speckled brown feathers} \\
3. \textcolor{blue}{some species mammals} \ul{do not have} \textcolor{orange}{speckled wings} \\
4. \textcolor{blue}{most species mammals} \ul{do not have} \textcolor{orange}{wings} \\

1. \textcolor{blue}{some birds} \ul{have} \textcolor{orange}{scales} \\
2. \textcolor{blue}{some birds} \ul{have} \textcolor{orange}{a speckled brown color} \\
3. \textcolor{blue}{some species mammals} \ul{have} \textcolor{orange}{wings} \\
4. \textcolor{blue}{most birds} \ul{don't have} \textcolor{orange}{wings} \\
5. \textcolor{blue}{most insects} \ul{have} \textcolor{orange}{wings} \\
6. \textcolor{blue}{most species mammals} \ul{don't have} \textcolor{orange}{wings} \\

\underline{\textbf{Target: most species mammals do not have wings}}
% some birds have a speckled brown color
% t:  0.09090909090909091
% some birds have scales
% t:  0.18181818181818182
% some birds are scaled
% t:  0.2727272727272727
% some birds are herbivores
% t:  0.36363636363636365
% some eagles live in open environments
% t:  0.45454545454545453
% some birds are eagles
% t:  0.5454545454545454
% some birds are herbivores
% t:  0.6363636363636364
% eagles live in an arctic environment
% t:  0.7272727272727273
% arctic animals live in an arctic environment

% some birds have a speckled brown color
% some birds have scales
% some birds have a speckled brown color
% some species mammals have wings
% most birds don't have wings
% most insects have wings
% most species mammals don't have wings
\end{tcolorbox}
\caption{Interpolation for T5VQVAE (top) and Optimus (bottom) where \textcolor{blue}{blue}, \ul{underline}, and \textcolor{orange}{orange} represent subject, verb, and object, respectively. Only unique sentences are shown.}
\label{tab:interpolation}
\end{table}

More importantly, we quantitatively evaluate the interpolation behaviour via the IS metric. We randomly select 100 (source, target) pairs and interpolate the path between them. Then, we calculate the averaged, maximal, and minimal ISs. As shown in Table \ref{tab:interpolation_smoothness}, T5VQVAE outperforms Optimus by over 43\% in average, which indicates that T5QVAE induces a latent space which can better separate the syntactic and semantic factors when contrasted to Optimus.
\begin{table}[t]
% \scriptsize
\setlength\tabcolsep{2.5pt}
\small
\centering
% \resizebox{7.8cm}{!}{
\renewcommand\arraystretch{1}
\begin{tabular}{llll}
\toprule
Evaluation Metrics & avg IS & max IS & min IS \\ \hline
Optimus(32, pretrain) & 0.22 & 0.53 & 0.13 \\ 
Optimus(768, pretrain) & 0.21 & 0.50 & 0.10 \\
T5VQVAE(base, soft) & \textcolor{blue}{\textbf{0.65}} & \textcolor{blue}{\textbf{1.00}} & \textcolor{blue}{\textbf{0.18}} \\ \toprule
\end{tabular}
% }
\caption{Interpolation smoothness.} \label{tab:interpolation_smoothness}
\end{table}

\paragraph{Latent Vector Arithmetics.} Inspired by word embedding arithmetics, e.g., $king - man + woman = queen$, we explore the compositional semantics via latent arithmetic with the target of sentence-level semantic control. After adding two latent vectors corresponding to two sentences $s_c = s_A + s_B$, we expect the resulting sentence to express the semantic information of both sentences. From Table \ref{tab:arithmetic_examples}, we can observe that T5VQVAE can generate the outputs containing both inputs' semantic information. E.g., the output contains \textit{are likely to} and \textit{their environment} from $s_A$ and \textit{to survive} and \textit{/} from $s_B$. In contrast, Optimus is not able to preserve to support this behaviour. Additional examples are provided in Appendix~\ref{sec:text_transfer_appendix} 
 (Table \ref{tab:appendix_arithmetic_more}). 
% traversals results
\begin{table}[t]
\centering
\begin{tcolorbox}[fontupper=\small, fontlower=\small] %, middle=0.3cm
% \begin{multicols}{2}
    \ul{\textbf{$s_A$: animals are likely to have the same color as their environment}}\\
    \ul{\textbf{$s_B$: animals require respiration to survive / use energy}} \\
    
    T5VQVAE: \textcolor{blue}{animals} \textcolor{orange}{are likely to} \textcolor{cyan}{survive / to survive} \textcolor{orange}{in their environment} \\
    Optimus: \textcolor{blue}{animals} have evolved from animals with traits that have an animal instinct
\end{tcolorbox}
\caption{Latent arithmetic $s_A + s_B$ for T5VQVAE(base) and Optimus(32). \textcolor{blue}{blue}, \textcolor{orange}{orange}, and \textcolor{cyan}{shallow blue} indicate the semantic information from both $s_A$ and $s_B$, from $s_A$ only, from $s_B$ only, respectively.}
\label{tab:arithmetic_examples}
\end{table}

\subsection{Inference Task}
Lastly, we move to downstream inference tasks, in which we aim to explore the controllability of T5VQVAE for reasoning with natural and symbolic languages. Specifically, we focus on two tasks including syllogistic-deductive natural language inference in EntailmentBank \cite{https://doi.org/10.48550/arxiv.2104.08661}, where a natural language conclusion has to be inferred from two premises, and mathematical expression derivation \cite{meadows2023symbolic}, where the goal is to predict the result of applying a mathematical operation to a given premise expression (written in latex).

\paragraph{Quantitative Evaluation.} We quantitatively evaluate several baselines following the same procedure as the AutoEncoding task. Table \ref{tab:metrics_inf} shows that T5VQVAE outperforms all VAE models on both benchmarks.
% \begin{center}
\begin{table}[t]
% \scriptsize
\setlength\tabcolsep{2.5pt}
\resizebox{7.8cm}{!}{
\small
\centering
\renewcommand\arraystretch{1}
\begin{tabular}{llllll}
\toprule
\multicolumn{6}{c}{\textit{Natural Language Inference (EntailmentBank)}} \\
Evaluation Metrics & BLEU & Cosine & BLEURT & Loss $\downarrow$ & PPL $\downarrow$ \\ \hline
T5(small) & 0.54 & 0.96 & 0.22 & 0.69 & 1.99 \\
T5(base) & \textbf{0.57} & \textbf{0.96} & \textbf{0.33} & \textbf{0.61} & \textbf{1.84} \\
Bart(base) & 0.54 & 0.96 & 0.17 & 0.63 & 1.87 \\
FlanT5(small) & 0.22 & 0.89 & -1.33 & 0.99 & 2.69 \\ 
FlanT5(base) & 0.32 & 0.89 & -0.31 & 0.95 & 2.58 \\ 
T5bottleneck(base) & 0.35 & 0.91 & -0.20 & 1.24 & 3.45 \\ \hdashline
Optimus(32) & 0.07 & 0.74 & -1.20 & 1.13 & 2.31 \\
Optimus(768) & 0.08 & 0.74 & -1.21 & 0.82 & 2.27 \\
DELLA(32) & 0.08 & 0.85 & -1.23 & 1.69 & 5.41 \\
DELLA(768) & 0.09 & 0.87 & -1.09 & 1.54 & 4.66 \\
T5VQVAE(small) & 0.11 & 0.73 & -1.23 & 0.85 & 2.33 \\
T5VQVAE(base) & \textcolor{blue}{\textbf{0.46}} & \textcolor{blue}{\textbf{0.94}} & \textcolor{blue}{\textbf{0.10}} & \textcolor{blue}{\textbf{0.84}} & \textcolor{blue}{\textbf{2.31}} \\ \hline \hline
\multicolumn{6}{c}{\textit{Mathematical Expression Derivation}} \\
Evaluation Datasets & EVAL & SWAP & EASY & EQ & LEN \\ \hline
T5(small) & 0.69 & 0.48 & 0.57 & 0.60 & 0.63 \\
T5(base)  & 0.97 & 0.65 & 0.90 & 0.72 & 0.81 \\ \hdashline
Optimus(32) & 0.72 & 0.50 & 0.59 & 0.23 & 0.40 \\
Optimus(768) & 0.79 & 0.56 & 0.63 & 0.29 & 0.44 \\
DELLA(32)  & 0.12 & 0.16 & 0.13 & 0.13 & 0.13 \\
DELLA(768) & 0.13 & 0.18 & 0.12 & 0.13 & 0.14 \\
T5VQVAE(small) & 0.75 & \textcolor{blue}{\textbf{0.57}} & 0.77 & \textcolor{blue}{\textbf{0.48}} & \textcolor{blue}{\textbf{0.50}} \\
T5VQVAE(base) & \textcolor{blue}{\textbf{0.76}} & 0.56 & \textcolor{blue}{\textbf{0.78}} & 0.47 & \textcolor{blue}{\textbf{0.50}} \\
\toprule
\end{tabular}
}
\caption{Quantitative evaluation on inference tasks.} \label{tab:metrics_inf}
\end{table}
% \end{center}

\paragraph{Qualitative Evaluation.} Next, we focus on the NLI task to explore the controllability of T5VQVAE for sentence-level inference traversing the latent space. 
As illustrated in Table \ref{tab:trav_inf_examples}, traversing the dimension corresponding to an individual word (e.g., \textit{object} from premise 1 (P1)) cannot preserve the target word during the traversal along with the semantic coherence of the transitions, indicating that the inference is done entirely in the Encoder. Therefore, we next explore how to manipulate the latent representation to deliver a more controllable inference behaviour. 
%the conclusion, which indicates that the inference process does not happen in the cross-attention module in Decoder. Instead, it happens inside the Encoder. The output of the Encoder contains the information about the conclusion directly, which works on the Decoder for guiding the generation of the conclusion.
% traversals results
\begin{table}[t]
\begin{tcolorbox}[fontupper=\small, fontlower=\small] %, middle=0.3cm
\underline{\textbf{P1: a human is a kind of \textcolor{blue}{object}}} \\
%\specialrule{0pt}{-1pt}{-1pt}
\underline{\textbf{P2: a child is a kind of young human}} \\
%\specialrule{0pt}{-1pt}{-1pt}
\underline{\textbf{C: a child is a kind of object}}\\ \\
dim6: a young object is a kind of child \\
dim6: a boy is a kind of young object \\
dim6: a little boy is a kind of young human
% dim6: a child is a kind of object \\
% dim6: a child is a kind of young human 
% dim6: a young human is a kind of object \\
% dim6: a baby is a kind of young object \\
% dim6: a young child is a kind of child
% dim 6 sent 9: a boy is a kind of young human 
% dim 6 sent 11: a child is a kind of young object 
% dim 6 sent 12: a baby is a kind of young human 
% dim 6 sent 13: a child is a kind of object 
% dim 6 sent 14: a child is an object that is human 
% dim 6 sent 15: a young object is a kind of child 
% dim 6 sent 16: a child is a kind of young person 
% dim 6 sent 17: a small boy is a kind of young human 
% dim 6 sent 18: a child is a kind of object 
% dim 6 sent 19: a child is a kind of young object
\end{tcolorbox}
\caption{T5VQVAE (base): traversed conclusions.}
\label{tab:trav_inf_examples}
\end{table}
% In addition, we explore whether the proposed discrete latent space can deliver a controllable symbolic-level inference behaviour. 

%\paragraph{Symbolic-level inference} 

Recent work \cite{zhang2023type} has provided a granular annotated dataset of step-wise explanatory inference types, which reflect symbolic (syllogistic-style) operations between premises and conclusions, including \textit{argument/verb substitution}, \textit{further specification}, and \textit{conjunction}.
We leverage this annotation to input two premises into the Encoder to derive the latent token embeddings of individual arguments and guide the generation of the conclusion via the Decoder. For example, for \textit{argument substitution} and \textit{verb substitution}, which refers to the process of obtaining a conclusion by substituting one argument/verb from the first premise to an argument/verb of the second premise, we substitute the respective token embeddings in the latent space and feed the resulting representation to the decoder. Table \ref{tab:quasi_1} shows that by substituting the embeddings of the arguments, we can control the behaviour of the model and elicit a systematic inference behaviour. We provide \textit{further specification} and \textit{conjunction} in Table \ref{tab:quasi_2}.
% traversals results
\begin{table}[t]
\begin{tcolorbox}[fontupper=\small, fontlower=\small]
P1: a \textcolor{blue}{shark} is a kind of \underline{fish} \\
P2: a \underline{fish} is a kind of aquatic animal \\
Pred: a \textcolor{blue}{shark} is a kind of aquatic animal
% P1: \textcolor{blue}{oxygen} is a kind of \underline{matter} \\
% P2: a vacuum has no \underline{matter} in it \\
% Pred: a vaccum has no \textcolor{blue}{oxygen} in it
% P1: \textcolor{blue}{heat} is a kind of \underline{energy} \\
% P2: flowing can be a kind of transfer of \underline{energy} \\
% Pred: flowing can be a kind of transfer of \textcolor{blue}{heat}
\tcblower
% P1: recyclable means a material can be \textcolor{blue}{recycled} / \underline{reused} many times \\
% P2: renewable resources can be \underline{used} over again \\
% Pred: renewable resources can be \textcolor{blue}{recycled} over again \\
P1: to \underline{move} something can mean to \textcolor{blue}{transfer} something \\
P2: flowing is a kind of \underline{movement} for energy \\
Pred: flowing is a kind of \textcolor{blue}{transfer} of energy
\end{tcolorbox}
\caption{T5VQVAE(base): quasi-symbolic inference examination in AutoEncoder (Top: argument substitution, Bottom: Verb substitution).}
\label{tab:quasi_1}
\end{table}
% The \textit{further specification}  type captures clausal-level sentence operations, where two premises build a conclusion with the introduction of a subordinate clause. Table \ref{tab:fur_or_con} shows that the model can deliver the expected symbolic (clausal substitution) behaviour. We emphasize that this was achieved by the direct manipulation of the latent space.  
% \input{tables/more_quasi_symbolic}
These results show that the latent embeddings can be manipulated to deliver a syllogistic-style inference behaviour. In particular, we demonstrate that the distributed semantic information in the latent space contains information about co-occurring tokens within the sentence that can be systematically localised (within specific arguments, predicates or clauses) and manipulated to generate a sound conclusion. This behaviour can be potentially leveraged as a foundation to build an interpretable and multi-step natural language inference model. More examples are reported in the Appendix \ref{sec:NLI_appendix}.

\section{Related work} \label{sec:related}
% \paragraph{Semantic control via latent space} \citet{zhang2022} reveal that the natural language explanations are clustered according to the content of the semantic role. In this case, its generation can be localised and controlled by manipulating the movement of its latent vector over different role-content clusters. Next, they explored the controllability of flow-based invertible neural network (INN) to autoencoder by learning the bijective transformation between the hidden space of autoencoder and multivariate Gaussian space of INN with the target of more separate role-content clusters \cite{zhang2023learning}. Before that, \citet{https://doi.org/10.48550/arxiv.2109.07169} manually defined the discrete latent space of language where each dimension controls different semantic information of a sentence. Instead of using one latent space, some works defined two or more latent spaces to control the generation of natural language. For example, \citet{bao-etal-2019-generating} defines semantic and syntactic latent spaces separately, allowing separate control over semantic or syntactic information during generation. Similar to them, \citet{john-etal-2019-disentangled} concentrated on the Text Style Transfer task. They defined two latent spaces - one for sentiment and another for semantics. Both of them use RNN-based VAE as baselines. In this work, we concentrated on the localised semantic control of the T5 via VQVAE with the target of delivering inference behaviour.
\paragraph{Semantic Control via Latent Spaces.} \citet{zhang2022quasi, zhang2023learning} investigated the semantic control of latent sentence spaces, demonstrating the basic geometric-semantic properties of VAE-based models. \citet{https://doi.org/10.48550/arxiv.2109.07169} defined disentangled latent spaces focusing on the separation between content and syntactic generative factors. Moreover, some works focused on defining two separate latent spaces to control natural language generation on specific downstream tasks, such as style-transfer and paraphrasing \cite{bao-etal-2019-generating, john-etal-2019-disentangled}. Comparatively, this work explores more granular control and a broader spectrum of tasks: from syllogistic to symbolic inference.

\paragraph{Language VAEs.} Instead of Optimus \cite{li2020optimus} and its variation \cite{fang-etal-2022-controlled, hu-etal-2022-fuse} where the encoder and decoder are BERT and GPT2, respectively, most of the language VAE literature are based on LSTM architectures instantiated on different text generation tasks, including story generation \cite{fang2021transformerbased}, dialogue generation \cite{zhao-etal-2017-learning}, text style transfer \cite{john-etal-2019-disentangled, shen2020educating}, text paraphrasing \cite{bao-etal-2019-generating}, among others. Some works also investigated different latent spaces or priors to improve representation capabilities \cite{dai-etal-2021-apo, ding-gimpel-2021-flowprior,fang-etal-2022-controlled}. Comparatively, this work contributes by focusing on the close integration between language models and vector-quantized VAE-driven granular control, instantiating it in the context of a state-of-the-art, accessible, and cross-task performing language model (T5).

% \paragraph{VQVAE in NLP} In contrast to the Image domain \cite{van2017neural, razavi2019generating}, the integration of vector-quantized VAEs is relatively less explored in natural language. \citet{https://doi.org/10.48550/arxiv.2110.05999} explored long text generation via VQVAE, where its discrete latent representations capture the global structure of the text. Before that, \citet{huang-ji-2020-semi} utilized a VQVAE-based model to perform semi-supervised event-type induction. Furthermore, \citet{https://doi.org/10.48550/arxiv.1905.12752} proposed the vector quantization as a residual network in the encoder of the transformer, which can perform paraphrasing after trained with an unlabeled monolingual corpus only. In contrast, this work focuses on addressing the question of whether vector-quantised VAEs can be used as a mechanism to deliver improved semantic control by bridging the gap between language models and VAE architectures. 

\section{Conclusion and Future Works} \label{sec:concl}   
In this work, we build a model for improving the semantic and inference control for VAE-enabled language model (autoencoding) architectures. We propose a new model (i.e., T5VQVAE) which is based on the close integration of a vector-quantized VAE and a consistently accessible and high-performing language model (T5). The proposed model was extensively evaluated with regard to its syntactic, semantic and inference controls using three downstream tasks (autoencoding, text transfer, and inference task). Our experimental results indicate that the T5VQVAE can outperform the canonical state-of-the-art models in those tasks and can deliver a quasi-symbolic behaviour in the inference task (via the direct manipulation of the latent space). 

As future work, we plan to further explore applications on symbolic natural language inference via the direct manipulation of the latent space, and to investigate the controllability of recent large language models through the VQVAE architecture. Moreover, additional research directions could be informed by the current work:

\paragraph{Word-level Disentanglement.} Our architecture provides a foundation to explore token/word-level disentanglement for more general sentence and inference representation tasks. While sentence-level disentanglement is widely explored in the NLP domain, such as sentiment-content \cite{john2019disentangled,hu2021causal}, semantic-syntax \cite{bao2019generating,zhang2023graph}, and negation-uncertainty \cite{vasilakes-etal-2022-learning}, or syntactic-level disentanglement \cite{felhi2022towards}, this mechanism is still under-explored in other NLP tasks \cite{liao-etal-2020-explaining}.

\paragraph{Interpretability.} Discrete properties derived from vector quantization can enable the further probing and interpretability of neural networks by discretizing continuous neural latent spaces, where symbolic concepts are emerging in both images \cite{deng2021discovering,li2023does} and natural language \cite{tamkin2023codebook} domains. 

\section*{Limitations} \label{sec:limit}
% While the proposed framework has delivered clearly observable properties in terms of generative control, understanding the uncertainty and safety guarantees of the model would require further investigation. 

While T5VQVAE can improve inference performance and deliver inference control on syllogistic-deductive style explanations, the application on more complex reasoning tasks (e.g. involving quantifiers and multi-hop inference) is not fully explored. Besides, we still observe limitations in out-of-distribution generalisation in the mathematical expressions corpus despite the improvement over existing VAE models in terms of robustness. This, in particular, is highlighted by the decrease in performance obtained on the length generalisation split (LEN) for both autoencoding and expression derivation tasks. 
%This work is limited to the T5 architecture, whether VQVAE can perform semantic control to recent large language models should be explored in the future.

% While our experiments show that T5VQVAE can significantly improve the performance of VAEs models on encoding mathematical expressions, 

\section*{Acknowledgements}
We appreciate the reviewers for their insightful comments and suggestions. This work was partially funded by the Swiss National Science Foundation (SNSF) project NeuMath (\href{https://data.snf.ch/grants/grant/204617}{200021\_204617}), by the EPSRC grant EP/T026995/1 entitled “EnnCore: End-to-End Conceptual Guarding of Neural Architectures” under Security for all in an AI enabled society, by the CRUK National Biomarker Centre, and supported by the Manchester Experimental Cancer Medicine Centre and the NIHR Manchester Biomedical Research Centre.

\bibliography{references}
\bibliographystyle{acl_natbib}

% \appendix
% \clearpage
\appendix
% \onecolumn
% \newpage
% \section{Overview}
% \begin{figure*}[ht!]
% \begin{center}
%     \includegraphics[scale=0.31]{figs/overview.png}
%     \caption{By controlling the discrete latent space, we can control the cross-attention to \textcolor{blue}{guide} the generation of the sentence generator. We focused on three progressively deeper tasks with the target of precise semantic control.}
%     \label{fig:overview}
%  \end{center}
% \end{figure*}
\section{Training setup} \label{sec:set_up}

\paragraph{Datasets} Table \ref{tab:stats_data} displays the statistical information of the datasets used in the experiment. As for the AutoEncoder setup, we use the non-repetitive explanations selected from both datasets as the experimental data. As for the Inference task, we use the data from EntailmentBank and Math Symbol Inference. The semantic roles of our data are annotated by automatic semantic role labelling tool \cite{Gardner2017ADS}.
% We report in Table~\ref{tab:srl_silva} the annotated categories and corresponding statistic information. The semantic roles of our data are annotated by automatic semantic role labelling tool \cite{Gardner2017ADS}.
\begin{table}[ht!]
    \small
    \centering
    \renewcommand\arraystretch{1}
      % \resizebox{7.6cm}{!}{
    \begin{tabular}{|c|cc|}
        \hline
        Corpus & Num data. & Avg. length \\ \hline
        WorldTree & 11430 & 8.65 \\
        EntailmentBank & 5134 & 10.35 \\ 
        Math Symbol & 32000 & 6.84 \\
        Math Symbol Inference & 32000 & 51.84 \\ \hline
        
    \end{tabular}
    % }
    \caption{Statistics from datasets.} \label{tab:stats_data}
\end{table}

\paragraph{T5VQVAE training} We use T5VQVAE(small) to choose the most appropriate codebook size between 2000 and 22000. In the experiment, the maximal epoch is 100. The learning rate is 5e-5. We use exponential moving averages (EMA) to update the codebook. Besides, we also investigated the optimal point within the architecture to learn the codebook. As shown in Table \ref{tab:diff_codebook}, T5VQVAE performs better when the codebook is learned at the end of the Encoder. This observation suggests that cross-attention is crucial in vector quantisation (VQ) learning.
% \begin{center}
\begin{table}[ht!]
\scriptsize
\setlength\tabcolsep{2.5pt}
\resizebox{7.8cm}{!}{
\small
\centering
\renewcommand\arraystretch{1}
\begin{tabular}{llllll}
\toprule
Metrics & BLEU & BLEURT & cosine & Loss $\downarrow$ & PPL $\downarrow$ \\ \hline
02000 & 0.73 & 0.21 & 0.93 & 0.79 & 2.20 \\
06000 & 0.79 & 0.45 & 0.95 & 0.61 & 1.84 \\
10000 & 0.81 & 0.62 & 0.97 & 0.46 & 1.58 \\
14000 & 0.82 & 0.62 & 0.96 & 0.42 & 1.52 \\
18000 & 0.83 & 0.64 & 0.96 & 0.38 & 1.46 \\
22000 & 0.83 & 0.67 & 0.96 & 0.34 & 1.40 \\ \hline \hline
\multicolumn{6}{c}{\textit{\textit{T5VQVAE(small) with different depth L in Encoder}}} \\ \hline
T5VQVAE(L=05) & 0.47 & -0.80 & 0.80 & 0.91 & 2.48 \\ 
T5VQVAE(L=04) & 0.59 & -0.56 & 0.84 & 0.76 & 2.13 \\ 
T5VQVAE(L=03) & 0.65 & -0.42 & 0.85 & 0.68 & 1.97 \\ 
% T5VQVAE(base, L=06) & 0.67 & -0.19 & 0.87 & 0.91 & - \\ 
T5VQVAE(L=02) & 0.70 & -0.21 & 0.88 & 0.65 & 1.91  \\ \toprule
\end{tabular}
}
\caption{T5VQVAE(small): Different sizes of codebook and optimal point.} \label{tab:diff_codebook}
\end{table}
% \end{center}

\paragraph{Expotential Moving Average (EMA)} Let $\{E(x_{k, 1}), ..., E(x_{k, n_k})\}$ be the set of word embedding $x_{k, i}$ belonging to the $z_k$. The optimal value for $z_k$ is the average of elements in this set, which can be described as:
\begin{align*}
z_k = \frac{1}{n_k}\sum_{i}^{n_k} E(x_i)
\end{align*}
However, we cannot use this to update $z_k$ since we usually work on mini-batches. Instead, we can use EMA to update $z_k$.
\begin{align*}
N_k^{(t)} &:= N_k^{(t-1)} \times \lambda + n_k^{(t)} (1-\lambda) \\
m_k^{(t)} &:= m_k^{(t-1)} \times \lambda + \sum_{i}  E(x_{k, i}) \\
z_k & := \frac{m_k^{(t)}}{N_k^{(t)}}
\end{align*}
Where $\lambda$ is 0.99 following the setup of \cite{van2017neural}.

\paragraph{Optimus and DELLA training setup} Both of them can be trained via the evidence lower bound (ELBO) on the log-likelihood of the data $x$ \cite{kingma2013auto}. To avoid KL vanishing issue, which refers to the Kullback-Leibler (KL) divergence term in the ELBO becomes very small or approaches zero, we select the cyclical schedule to increase weights of KL $\beta$ from 0 to 1 \cite{fu-etal-2019-cyclical} and KL thresholding scheme \cite{li-etal-2019-surprisingly} that chooses the maximal between KL and threshold $\lambda$. The final objective function can be described as follows:
\begin{align*} \label{eq:elbo_loss}
\mathcal{L}_\text{VAE} = & \mathbb{E}_{q_\phi(z|x)} \Big[ \log p_{\theta} ( x | z ) \Big]  \\
&- \beta \max \left[ \lambda , \text{KL} q_\phi(z|x) || p(z) \right ]
\end{align*}
\begin{center}
\begin{table*}[ht!]
\scriptsize
\setlength\tabcolsep{3pt}
\small
\centering
% \resizebox{7.8cm}{!}{
\renewcommand\arraystretch{1}
\begin{tabular}{p{6cm}p{6cm}p{1.5cm}p{1.5cm}}
\toprule
Golden Explanations & Predicted Explanations & BLEURT & BLEU \\ \hline
the grand canyon is a kind of place & the grand canyon is a kind of canyon & 0.26 & 0.87 \\ \hline
a blood thinner can be used to treat people with heart attacks and strokes & a heart thinner can be used to treat people with blood and heart & -0.05 & 0.44 \\ \hline
the plant offspring has yellow flowers & offspring means offspring & -1.30 & 0.12 \\ \hline
lack is similar to ( low ; little ) & little means ( little ; little ) in quality & -1.18 & 0.44 \\ \hline
preserved means ( from the past ; from long ago ) & preserved means used to be ( preserved ; preserved ) from a long time & -0.01 & 0.50 \\ \hline
the plant offspring has yellow flowers & offspring means offspring & -1.30 & 0.12 \\ \hline
electricity causes less pollution than gasoline & gasoline causes less gasoline than gasoline & -0.22 & 0.66 \\ \hline
insulin is a kind of hormone & insulin is made of insulin & -0.31 & 0.49 \\ \hline
living things all require a producers for survival & living things all require a living thing for survival & 0.03 & 0.77 \\ \hline
gravity causes nebulas to collapse & gravity causes a sleef of an artery to collapse & -1.30 & 0.44 \\ \hline
out is synonymous with outside & outward is synonymous with out & -0.36 & 0.80 \\ \hline
to prevent means to make it not happen & to make means to not happen & -0.74 & 0.71 \\ \hline
a branch is a kind of object & a branch is a kind of branch & -0.03 & 0.85 \\ \hline
force requires energy & force means amount & -0.40 & 0.33 \\ \hline
spot means location & place means kind of place & -0.14 & 0.20 \\ \hline
gritty is similar to rough & grease is similar to grease & -0.80 & 0.60 \\ \hline
sidewalk means pavement & bike means bike & -0.62 & 0.33 \\ \hline
a gravel pit is a kind of environment & a gravel pit is a kind of gravel & 0.03 & 0.87 \\ \hline
a electron has a negative ( -1 ) electric charge & a electron has a negative ( electric charge ; negative charge ) & 0.23 & 0.75 \\ \hline
fish is a kind of meat & fish are a kind of fish & -0.29 & 0.66 \\
jogging is similar to running & running is a kind of running & -0.23 & 0.33 \\ \hline
the speed of the sailboat can be calculated by dividing 35 by 5 & the speed of the boat can be calculated by dividing the length of a boat & 0.20 & 0.60 \\ \hline
if an object has 0 mechanical energy then the object will stop moving & if an object has a mechanical energy then the object has to move to 0 & 0.09 & 0.66 \\
\toprule
\end{tabular}
% }
\caption{T5VQVAE(base): more examples with low BLEURT score.} \label{tab:more_example_1}
\end{table*}
\end{center}

\section{Visualization}
\begin{figure}[ht!] 
    \centering
    \includegraphics[width=\columnwidth]{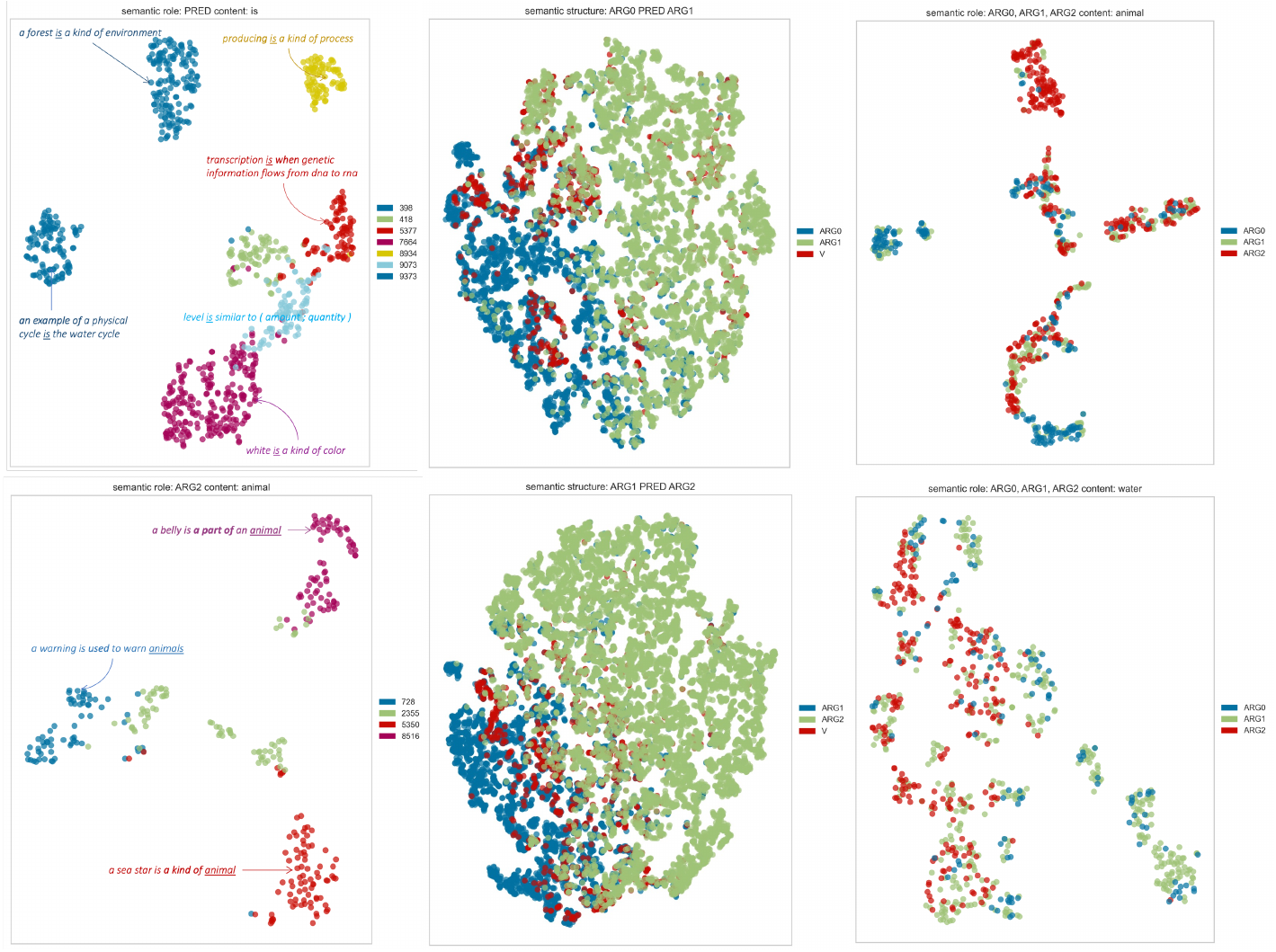} 
    \caption{t-SNE plot of the T5VQVAE latent space. Left: same role-content(\textit{PRED-is}, \textit{ARG2-animal}). Middle: different role-content(\textit{ARG0-PRED-ARG1}, \textit{ARG1-PRED-ARG2}). Right: different roles with same content (\textit{ARG0, 1, 2 - animal}, \textit{ARG0, 1, 2 - water}).}
    \label{fig:tsne_big} 
\end{figure}

In Figure \ref{fig:tsne_big}, we visualise the latent space of T5VQVAE via t-distributed Stochastic Neighbor Embedding (T-SNE) \cite{van2008visualizing} to analyse the organization of key semantic clusters. Specifically, we visualize the clusters of token embeddings with the same role-content, different roles, and the same content with different roles, respectively. We can observe that under the same role-content (left), the latent token embeddings are widely distributed in the latent space as the representation of the role-content is affected by the context, which indicates poor disentanglement. For different roles (middle), there are big overlaps between different semantic roles, which indicates poor disentanglement of semantic role structure. For the same content with different roles (right), it can be observed that different semantic role clusters are fully overlapped. Those visualizations indicate that the semantic information is naturally entangled after an attention-based Encoder. 

\section{AutoEncoding Task} \label{sec:autoencode_appendix}
We provide more reconstructed explanations with low BLEURT scores in Table \ref{tab:more_example_1}. we manually evaluate its performance and show the common issues in the AutoEncoding setup. (1) repetition: some explanations that describe the synonym are suffered from information loss. E.g., the prediction is \textit{the grand canyon is a kind of canyon} where the golden is \textit{the grand canyon is a kind of place}. (2) wrong numerical token: the model cannot precisely reconstruct the numerical token. E.g., \textit{the speed of the boat can be calculated by dividing the length of a boat} compared with the golden: \textit{the speed of the sailboat can be calculated by dividing 35 by 5}.

\section{Text Transfer Task} \label{sec:text_transfer_appendix}
We provide more traversal, interpolation, and arithmetic examples in Tables \ref{tab:appendix_traversal},\ref{tab:appendix_traversal_more}, \ref{tab:appendix_traversal_more_1}, and \ref{tab:appendix_arithmetic_more}.
% traversals results
\begin{table*}[ht!]
\begin{tcolorbox}[fontupper=\small, fontlower=\small, title=Traversal]
\textbf{\underline{an animal requires warmth in cold environments}} \\
\begin{multicols}{2}

dim0: animals usually maintain a safe distance from predators during the hibernation process \\
dim0: animals usually require warmth in cold temperatures for survival \\
dim0: animals must sense prey to survive / find food \\
dim0: animals must sense food to survive in the cold environment \\

dim1: animals must protect themselves ( against predators ; from predators ) \\
dim1: animals with pacemakers must sense danger in order to eat prey \\
dim1: animals with sensory organs provided shelter in cold environments \\
dim1: animals with diabetes should be protected from predators in the water \\

dim2: animals must sense ( predators ; food ) to survive \\
dim2: animals must sense other animals for food / shelter \\
dim2: animals must sense other animals for survival in cold environments \\
dim2: animals with circulatory system have a positive impact on themselves by breathing air \\

\columnbreak
%\\ \\ \\

dim4: animals with cold cardiovascular systems can survive in cold environments by breathing \\
dim4: animals must sense prey to survive in cold environments \\
dim4: animals must sense other animals for survival while they are at sea; in an environment \\
dim4: animals usually nurse their offspring through the winter \\

dim5: animals must sense prey to survive and reproduce \\
dim5: animals must sense food to find food \\
dim5: animals must sense prey in order to survive survival in the cold environment \\
dim5: animals require warmth in cold environments to ( survive ; find food ) \\

dim6: animals must sense food in order to survive in cold environments \\
dim6: animals must sense prey in order to survive / find food \\
dim6: animals with heat - circulatory system must cool themselves in cold environments \\
dim6: animals must sense prey to survive in cold environments \\
\end{multicols}
\end{tcolorbox}
\caption{Traversal for Optimus latent space.}
\label{tab:appendix_traversal}
\end{table*}
% traversals results
\begin{table*}[ht!]
\begin{tcolorbox}[fontupper=\small, fontlower=\small, title=Traversal]
\textbf{\underline{an astronaut requires the oxygen in a spacesuit backpack to breathe}} \\
\begin{multicols}{2}

dim1: an \textcolor{blue}{astronaut} requires the oxygen in a spacesuit backpack to breathe \\
dim1: an \textcolor{blue}{organism} requires the oxygen in a spacesuit backpack to breathe \\
dim1: an \textcolor{blue}{animal} requires the oxygen in a spacesuit backpack to breathe \\
dim1: an \textcolor{blue}{student} requires the oxygen in a spacesuit backpack to breathe \\

dim2: an astronaut \textcolor{blue}{requires} the oxygen in a spacesuit backpack to breathe \\
dim2: an astronaut \textcolor{blue}{can wear} the oxygen in a spacesuit backpack to breathe \\
dim2: an astronaut \textcolor{blue}{requires} the oxygen in a spacesuit backpack to breathe \\
dim2: an astronaut \textcolor{blue}{requires} the oxygen in a spacesuit backpack to breathe \\

\columnbreak
%\\ \\ \\

dim1: astronauts wear spacesuits in the space station to avoid the issue of heat loss after a space probe \\
dim1: astronauts wear spacesuits in the space environment to protect the astronaut from harmful chemical reactions \\
dim1: astronauts wear spacesuits in the space station to keep the body warm \\
dim1: astronauts wear spacesuits in the spacesuit worn by the astronauts to take in oxygen \\

dim2: astronauts wear spacesuits in the space station in space \\
dim2: astronauts conducting the orbit of the moon in space during the last stage of a lunar cell might cause direct sunlight to lands on the moon \\
dim2: astronauts wear on the body the oxygen in a spacesuit backpack after the spacecraft escapes the atmosphere \\
dim2: astronauts wear spacesuits in the space station to protect the body of an astronaut \\

\end{multicols}
\end{tcolorbox}
\caption{Traversal comparison (left: T5VQVAE(base), right: Optimus).}
\label{tab:appendix_traversal_more}
\end{table*}
% traversals results
\begin{table}[ht!]
\begin{tcolorbox}[fontupper=\small, fontlower=\small, title=Traversal]
\textbf{\underline{pedals are a kind of object}} \\
% \begin{multicols}{2}
dim0: \textcolor{blue}{pedals} are a kind of pedal \\
dim0: \textcolor{blue}{pedaling} is a kind of object \\
dim0: \textcolor{blue}{a pedal} is a kind of object \\
dim0: \textcolor{blue}{leather} is a kind of object \\

dim1: a pedal \textcolor{blue}{is} a kind of object \\
dim1: pedals \textcolor{blue}{are} a kind of object \\
dim1: pedals \textcolor{blue}{are} a kind of object \\
dim1: a pedal \textcolor{blue}{is} a kind of object \\

% \columnbreak
%\\ \\ \\

dim0: objects are a kind of kind of nonliving thing \\
dim0: rust is a kind of object \\
dim0: objects are a kind of kind of heavy object \\
dim0: rust is a kind of object \\

dim1: objects are a kind of kind of nonliving thing \\
dim1: rust is a kind of object \\
dim1: bones are a kind of object \\
dim1: objects are a kind of kind of small particle \\
% \end{multicols}

\textbf{\underline{travel means to move}} \\
% \begin{multicols}{2}

dim2: travel \textcolor{blue}{means} move \\
dim2: travel \textcolor{blue}{is similar} to move \\
dim2: travel \textcolor{blue}{is used} to move \\
dim2: travel \textcolor{blue}{is a kind of} movement \\

dim3: travel means \textcolor{blue}{to move} \\
dim3: travel means \textcolor{blue}{stay} \\
dim3: travel means to \textcolor{blue}{withstand travel} \\ 
dim3: travel means to \textcolor{blue}{be transported} \\

% \columnbreak
%\\ \\ \\

dim2: to move means to move \\
dim2: to pedal means to move something faster \\
dim2: to move means to move \\
dim2: to move means to move \\

dim3: to raise means to move something \\
dim3: to pedal means to move faster \\
dim3: to move means to move \\
dim3: to pedal means to move quickly
% \end{multicols}
\end{tcolorbox}
\caption{Traversal comparison (top: T5VQVAE(base), bottom: Optimus). We can observe that T5VQVAE can provide better semantic control than Optimus.}
\label{tab:appendix_traversal_more_1}
\end{table}
\begin{table}[ht!]
\begin{tcolorbox}[fontupper=\small, fontlower=\small, title=Arithmetic]
\textbf{\underline{$x_A$: a forest is a kind of land}}  \textbf{\underline{$x_B$: a tornado is narrow in width}} \\

T5VQVAE: a tornado is small in land \\
Optimus: plants are a kind of resource \\

\textbf{\ul{$x_A$: a rabbit is a kind of animal that may live in a meadow}}  \textbf{\ul{$x_B$: december is during the winter in the northern hemisphere}} \\

T5VQVAE: december is a kind of animal that may be in a winter \\
Optimus: a animal can usually find something to eat \\

\textbf{\ul{$x_A$: fossil fuels are formed from dead prehistoric organisms}}  \textbf{\ul{$x_B$: orange is a kind of color}} \\

T5VQVAE: orange fossil fuels are formed from dead prey \\
Optimus: prehistoric organisms developed defenses against disease by compacting and burying large amounts of remains \\

\textbf{\ul{$x_A$: waves travel outward from the source}}  \textbf{\ul{$x_B$: water is made of matter}} \\

T5VQVAE: water points away from the source \\
Optimus: transverse waves cause the person to move perpendicular to the direction of the wave \\

\textbf{\ul{$x_A$: rotation is a kind of motion}}  \textbf{\ul{$x_B$: Leo is a kind of constellation}} \\

T5VQVAE: Leo is a kind of motion \\
Optimus: friction occurs when two object colliding causes the speed of their movement to increase \\

\textbf{\ul{$x_A$: the milky way is a kind of galaxy}}  \textbf{\ul{$x_B$: a rock is usually a solid}} \\

T5VQVAE: the milky way is usually a solid \\
Optimus: x -sex cells are inherited characteristics
\end{tcolorbox}
\caption{Addition Arithmetic comparison. We can observe that the T5VQVAE can hold the semantic information of both sentences after addition arithmetic.}
\label{tab:appendix_arithmetic_more}
\end{table}

\section{Inference Task} \label{sec:NLI_appendix}
We provide more examples in Tables \ref{tab:quasi} and \ref{tab:quasi_2}.
\begin{table}[ht!]
\begin{tcolorbox}[fontupper=\small, fontlower=\small, title=Argument and Verb substitution]
P1: \textcolor{blue}{heat} is a kind of \underline{energy} \\
P2: flowing can be a kind of transfer of \underline{energy} \\
Pred: flowing can be a kind of transfer of \textcolor{blue}{heat} \\

P1: \underline{aluminum} is always \textcolor{blue}{nonmagnetic} \\
P2: aluminum cans are made of \underline{aluminums} \\
Pred: aluminum cans are \textcolor{blue}{nonmagnetic} \\

P1: \textcolor{blue}{ground water} is a kind of \underline{water} \\
P2: a desert environment is low in availability of \underline{water} / availability of food \\
Pred: a desert environment is low in availability of \textcolor{blue}{ground water} / availability of food \\

P1: \underline{marine fossils} are \textcolor{blue}{fossils of water animals} \\
P2: \underline{marine fossils} are found in mountains \\
Pred: marine fossils are fossils of marine animals \\
Gold: fossils of water animals are found in mountains
\tcblower
P1: in order to \underline{breath} air you must \textcolor{blue}{take in} air \\
P2: a lung is a kind of organ for \underline{breath} air \\
Pred: a lung is a kind of organ for \textcolor{blue}{taking in} air \\

P1: \textcolor{blue}{running} is a kind of \underline{movement} \\
P2: an animal requires energy to \underline{move} \\
Pred: an animal requires energy to \textcolor{blue}{run} \\

P1: \underline{eating} is a kind of method for \textcolor{blue}{consuming} \\
P2: decomposers obtain nutrients by \underline{eating} waste \\
Pred: decomposers obtain nutrients by \textcolor{blue}{consuming} waste \\

P1: \underline{rolling} on a surface is a kind of \textcolor{blue}{moving} on a surface \\
P2: the rubber ball is \underline{rolling} across a surface \\
Pred: the rubber ball is \textcolor{blue}{moving} across a surface 
\end{tcolorbox}
\caption{T5VQVAE(base): quasi-symbolic inference examination in AutoEncoder (Top: argument substitution, Bottom: Verb substitution).}
\label{tab:quasi}
\end{table}
\begin{table}[ht!]
\begin{tcolorbox}[fontupper=\small, fontlower=\small, title=Further specification and Conjunction]
P1: a plant requires energy \textcolor{blue}{to make food} \\
P2: plants get energy from sunlight \\
Pred: plants get energy from sunlight \textcolor{blue}{to make food} \\

P1: a mountain is \textcolor{blue}{made of rocks} \\
P2: a mountain range contains mountains \\
Pred: a mountain range contains mountains \textcolor{blue}{made of rocks} \\

P1: a dinosaur is a kind of \textcolor{blue}{animal} \\
P2: dinosaurs are cold-blooded \\
Pred: dinosaurs are cold - blooded \textcolor{blue}{animals} \\

P1: \textcolor{blue}{gravity causes} orbits \\
P2: planets in the solar system orbit the sun \\
Pred: gravity causes planets in the solar system to orbit the sun \\

P1: scattering light \textcolor{blue}{decreases visibility} \\
P2: water droplets scatter light \\
Pred: water droplets scattering light \textcolor{red}{decreases air droplets} \\
Gold: water droplets scattering light decreases the visibility
\tcblower
P1: sound travels \textcolor{blue}{slowest through gas} \\
P2: sound travels \textcolor{blue}{fastest through solid} \\
Pred: sound waves travel fastest \textcolor{red}{through gas and gas} \\
Gold: sound travels fastest through solid and slowest through gas \\

P1: matter in the solid phase has definite \textcolor{blue}{volume} \\
P2: matter in the solid phase has definite \textcolor{blue}{shape} \\
Pred: matter in the solid phase has definite shape and volume \\

P1: a plant requires \textcolor{blue}{nutrients} from soil to grow \\
P2: a plant requires \textcolor{blue}{water} from soil to grow \\
Pred: a plant requires water and nutrients from soil to grow \\

P1: a hurricane has \textcolor{blue}{high wind speed} \\
P2: a hurricane has \textcolor{blue}{large amount of rain} \\
Pred: a hurricane has high wind speed and large amount of rain \\

P1: fungi \textcolor{blue}{can be multicellular} \\
P2: fungi \textcolor{blue}{have no chlorophyll} \\
Pred: fungi have no chlorophyll and can be multicellular \\
\end{tcolorbox}
\caption{T5VQVAE(base): quasi-symbolic inference examination in AutoEncoder (Top: further specification, Bottom: conjunction).}
\label{tab:quasi_2}
\end{table}

% \section{Explanation Semantic Roles} \label{sec:dsr_labels}
% We report in Table~\ref{tab:srl_silva} the annotated categories and corresponding statistic information.

% \input{tables/silva_srl.tex}

% % \input{tables/stats_role.tex}

% \newpage
% \input{sections/metrics.tex}

% \newpage
% \section{Unsupervised INN: explanation reconstruction} \label{sec:un_rec_example}

% Table \ref{tab:unsup_rec_explain} shows some generated explanations from AutoEncoder and unsupervised INN. As we can seen, they can reconstruct the explanations with good quality.
% \input{tables/rec_examples1}

% \newpage
% \section{Supervised INN: Explanation reconstruction} \label{sec:rec_example}

% Table \ref{tab:rec_explain} shows some reconstructed explanations from AutoEncoder, unsupervised INN, and supervised INN, respectively.
% \input{tables/rec_examples}

% % \section{Interpolation}

% % \input{tables/interpolation_examples_appendix}

\end{document}